\def\year{2022}\relax
\documentclass[letterpaper]{article} 
\pdfoutput=1
\usepackage{aaai22}  
\usepackage{times}  
\usepackage{helvet}  
\usepackage{courier}  
\usepackage[hyphens]{url}  
\usepackage{graphicx} 
\usepackage{booktabs}
\usepackage{subfigure}
\usepackage{natbib}  
\usepackage{caption} 
\DeclareCaptionStyle{ruled}{labelfont=normalfont,labelsep=colon,strut=off} 
\usepackage{amsmath,amssymb,amsfonts}
\usepackage{algorithm}
\usepackage{algorithmic}

\usepackage{newfloat}
\usepackage{listings}
\floatstyle{ruled}
\newfloat{listing}{tb}{lst}{}
\floatname{listing}{Listing}
\title{Learngene: From Open-World to Your Learning Task}
\author{
    Qiufeng Wang\equalcontrib,
    Xin Geng\thanks{Corresponding author.},
    Shuxia Lin\equalcontrib, 
    Shiyu Xia\equalcontrib, 
    Lei Qi, 
    Ning Xu
}

\affiliations{

    MOE Key Laboratory of Computer Network and Information Integration, China\\
    School of Computer Science and Engineering, Southeast University, Nanjing 210096, China\\
    \{qfwang, xgeng, shuxialin, shiyu\_xia, qilei, xning\}@seu.edu.cn

}

\usepackage{bibentry}

\begin{document}

\maketitle

\begin{abstract}
Although deep learning has made significant progress on fixed large-scale datasets, it typically encounters challenges regarding improperly detecting unknown/unseen classes in the open-world scenario, over-parametrized, and overfitting small samples. Since biological systems can overcome the above difficulties very well, individuals inherit an innate gene from collective creatures that have evolved over hundreds of millions of years and then learn new skills through few examples. Inspired by this, we propose a practical collective-individual paradigm where an evolution (expandable) network is trained on sequential tasks and then recognize unknown classes in real-world. Moreover, the learngene, i.e., the gene for learning initialization rules of the target model, is proposed to inherit the meta-knowledge from the collective model and reconstruct a lightweight individual model on the target task. Particularly, a novel criterion is proposed to discover learngene in the collective model, according to the gradient information. Finally, the individual model is trained only with few samples on the target learning tasks. We demonstrate the effectiveness of our approach in an extensive empirical study and theoretical analysis.
\end{abstract}

 \begin{figure}[t]
\centering
\centerline{\includegraphics[width=0.5\textwidth]{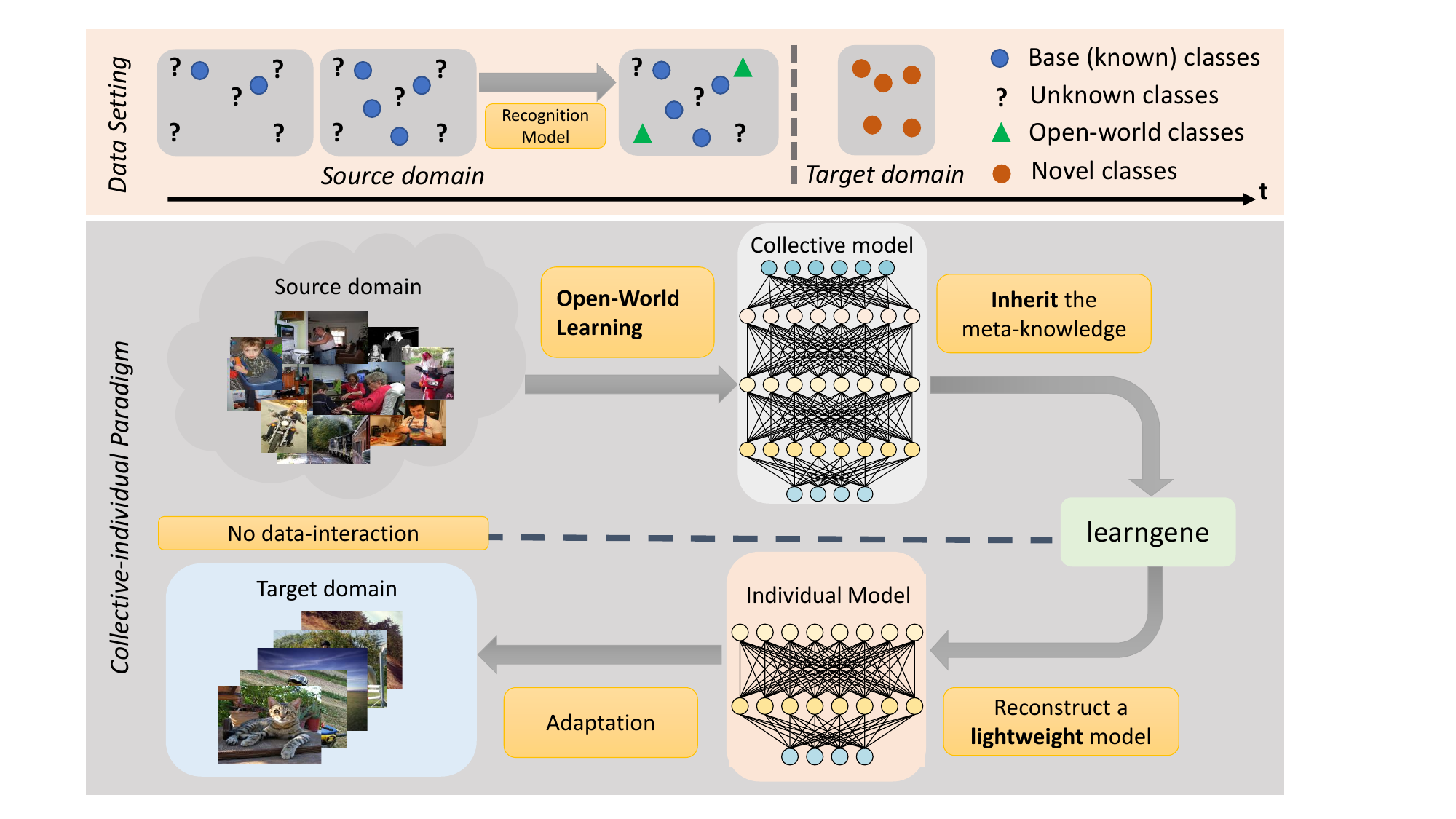}}
\caption{\emph{Approach Overview}: \emph{Top row}: The collective model is trained on continual learning tasks composed of base classes (as blue circles), and then recognize unknown classes (denoted by ‘?’) as open-world classes (green triangles, not belong to base classes), which are added to the training data incrementally. Finally, the individual model that inherits the meta-knowledge from the collective model adapts to the target tasks based on novel classes. \emph{Bottom row}: The expandable collective model continuously deals with continual learning tasks and then recognize open-world classes. The lightweight individual model is initialized and reconstructed with learngene. The individual model adapts quickly to the target learning task even in the absence of the source data (\textbf{no data-interaction}). }
\label{fig:paradigm} 
 \vskip -0.2in
\end{figure}

\section{Introduction}
Over the past decade, deep learning has made significant progress on a fixed large-scale dataset or in a closed-world, where only the instances seen during training are presented to them. However, a network, trained in a specific task, is incapable of recognizing unknown/unseen classes due to domain shift  \cite{ben2007analysis} and category shift \cite{xu2018deep}. Consequently, static and dedicated neural networks cannot resolve incrementally learning the unknown/unseen classes   \cite{DBLP:conf/kdd/FeiW016,DBLP:conf/www/XuLSY19} without retraining from scratch and without catastrophic forgetting \cite{french1999catastrophic}, which is exactly what \emph{open-world learning} \cite{DBLP:conf/cvpr/BendaleB15, DBLP:journals/corr/abs-1801-05609} wants to solve. In addition, neural networks are typically over-parametrized \cite{DBLP:journals/corr/HanMD15} and easily overfitting few samples \cite{wang2020generalizing}. 

 There are currently some learning paradigms that can solve some of the above problems, such as \emph{continual learning} \cite{hadsell_embracing_2020}, \emph{transfer learning} \cite{perkins1992transfer} and \emph{meta-learning} \cite{DBLP:journals/corr/abs-2004-05439}. Although continual learning can alleviate catastrophic forgetting by training on sequential tasks, the traditional continual learning cannot be reproduced in few samples. Many algorithms deal with category shift in transfer learning, which can reuse the model for the new task. In addition, meta-learning aims to design the model for learning prior knowledge of model parameters, so that new knowledge can be learned well from few samples. However, transfer learning comes across the dilemma of limited tasks, because it only focuses on the source tasks that transfer knowledge to the target tasks. Nevertheless, transfer learning and meta-learning perform inferior on the open-world tasks, as we discuss in Related Work section

Consequently, the above learning mechanisms are all limited. When we recall the learning process of biology, most of the behaviors of animals are congenital, not entirely through acquired learning. Besides, a recent neuroscience commentary, “What \emph{artificial neural networks} (ANNs) can learn from animal brains” \cite{zador2019critique} provides a critique of how rapid learning is currently implemented in ANNs. The innate knowledge comes from the universal \emph{genes} \cite{zador2019critique} of collective creatures that are constantly evolving. Afterwards, genes with certain rules flow to instantiated individuals, which enables them to \emph{learn} very rapidly. Inspired by this, this paper is strongly motivated towards these goals of innate knowledge and learning. Specifically, the enlightenment for ANNs is that it is possible to design the expandable neural network \textbf{(collective model)} in the outer loop to train sequential tasks in the open-world, and then 
discover the critical modules through a judgment criterion of gradient information. The critical modules named \textbf{learngene} inherit meta-knowledge from the collective model. The lightweight network \textbf{(individual model)} is initialized and reconstructed from learngene. Finally, the individual model is retrained with few samples and adapt quickly to the novel classes without data- interaction of the collective model.

To verify this premise, we turn to training and analysis of expandable neural networks. As shown in Figure~\ref{fig:paradigm}, we leverage the basic \emph{convolutional neural network} (CNN) to set up the collective model, which automatically expands the network with different sequence tasks of base classes. The pretrained collective model recognizes unknown classes as open-world classes or base classes and incrementally adds them to training set.
Afterwards, we propose the judgement criterion that the last/top $k$ network layers cover the meta-knowledge according to the changes of each layer of network parameters over time. Therefore, the $topk$ layers of the model are selected as the base modules of learngene. In other words, our paradigm replaces data-interaction with model interaction, thereby protecting the privacy of data. Besides, we employ the \emph{elastic weight consolidation} (EWC) \cite{kirkpatrick2017overcoming} loss to retain the innate generalization ability as much as possible. Finally, through adjoining a small feature layer in front of the base module and thinner fully connected layers in the back, the individual model evaluates the learning ability on the target data of novel classes. 

\textbf{Contributions.}  (1) We propose a novel collective-individual paradigm, which is applicable for the practical scenario of continual open-world classification to reduce the sample and capacity cost of institutions or individuals without data-interaction. (2) To realize the collective-individual paradigm, we present learngene that inherits the meta-knowledge from the collective model and reconstructs a lightweight model for the target task. (3) We propose an effective criterion that can discover the learngene in the collective model, according to gradient information. (4) We demonstrate the effectiveness of our approach in an extensive empirical study and theoretical analysis.

\section{Problem Setup}

In this section, we describe the collective-individual paradigm and learngene that inherits the meta-knowledge from the collective model in the open-world and reconstructs a  lightweight individual model for the target task.

\subsection{Preliminaries}
\textbf{Notation.} Let $\mathcal{X}$ and $\mathcal{Y}$ be a sample space and a label space, respectively. We refer to a pair of $\mathcal{Z}:=\mathcal{X} \times \mathcal{Y}$. In addition, let $\mathcal{H}=\{h: \mathcal{X} \rightarrow \mathcal{Y}\}$ be a hypothesis space and $\ell: \mathcal{Y} \times \mathcal{Y} \rightarrow \mathbb{R}^{+}$ represents a loss function. At any time $t$, we consider the set of known base classes as
$\mathcal{C}^{t}=\{1,2, . ., \mathrm{N}\}$ for the collective model. In order to realistically model the dynamics of the real-world case, we also assume that there exists a set of open-world classes $\mathcal{U}=\{\mathrm{N}+1, \ldots\}$, which could be encountered during inference. In addition, novel classes $\mathcal{N}=\{1, \ldots, K\}$ are different from previous classes, sampling for the reconstructed individual model. 

\textbf{Performance Measure.} The fundamental purpose of the collective-individual model is to maximize the performance on the target task of novel classes. Therefore, a standard procedure is to empirically estimate by empirical risk estimator, i.e., $\widehat{er}(h):=\frac{1}{n} \sum_{j=1}^{n} \ell(h, \mathcal{N})$, where $n$ is number of the target tasks and $h$ is the individual model.

\begin{table*}[htb]
\setlength{\abovecaptionskip}{0pt}
\setlength{\belowcaptionskip}{10pt}
\caption{Comparison of different task settings of collective-individual paradigm.}
\label{tab:related-work}
\vskip -0.4in
\begin{center}
\begin{scriptsize}
\begin{sc}
\begin{tabular}{clcccc}
\hline
\multicolumn{2}{c}{\upshape{Task setting}}                       &  {\upshape No data-interaction} & \upshape{Open-world} & \upshape{Inheriting model to reconstruct lightweight model} & \upshape{Few samples adaptation} \\ \hline
\multicolumn{2}{c}{\upshape{Transfer learning}}              & $\surd$                & $\times$          & $\times$                      & $\times$                   \\
\multicolumn{2}{c}{\upshape{Continual learning}}             & $\times$                & $\surd$          & $\times$                      & $\times$                   \\
\multicolumn{2}{c}{\upshape{Few-shot learning}}                  & $\surd$                & $\times$          & $\times$                      & $\surd$                   \\ \hline
\multicolumn{2}{c}{\upshape{Collective-individual paradigm}} & $\surd$                & $\surd$          & $\surd$                      & $\surd$                   \\ \hline
\end{tabular}
\end{sc}
\end{scriptsize}
\end{center}
\vskip -0.2in
\end{table*}

\subsection{The Collective-individual Paradigm}
\subsubsection{Definition 1:}
(\emph{collective-individual paradigm})
\emph{The collective model can discriminate a test instance belonging to the known $\mathrm{N}$ classes, and also recognize an unknown or unseen instance by classifying it as an open-world or base class. Afterwards, the collective model incrementally adds $\mathrm{M}$ new classes and updates itself to produce a predicted model $\mathcal{M}_\mathrm{N+M}$ without retraining from scratch on the whole dataset. The known class set is updated $\mathcal{C}_{t+1}=\mathcal{C}_{t}+\{\mathrm{N+1}, \ldots, \mathrm{N+M}\}$. In the collective-individual paradigm, a factory manufactures the collective model with base classes and open-world classes. The micro-institution or user trains the individual model initialized from learngene on the target learning task of novel classes.}

\subsubsection{Definition 2:}
(\emph{Learngene})
\emph{We formally define learngene as a rule $\phi$ for initializing the target model, inspired by \cite{zador2019critique}. Here we use gradient optimization rules to inherit important modules and parameters from the collective model and initialize the individual model. }

Our paradigm has the following characteristics:
i) \textbf{No data-interaction: }The individual model adapts to the target task in the absence of the source data, thereby protecting the privacy of data. Specifically, we allow category shift between the collective model and the individual model;
ii) \textbf{Open-world: }The collective model adjusts scale to adapt to new tasks continuously and adds the unknown/unseen classes incrementally. The model preserves knowledge of previous tasks, i.e., factories or platforms produce large-scale model;
iii) \textbf{Inheriting model to reconstruct lightweight model: } Inherit important modules and parameters from the collective model and initialize individual model;
iv) \textbf{few samples adaptation: }Micro-institutions or users could train the individual model only with few samples on target learning tasks. We distinguish between the collective and individual model by adding a subscript \emph{col} or \emph{idu} to each notation introduced above (e.g., $\theta_{\mathrm{col}}, \mathcal{D}^{idu}$). 

\section{Related Work} 
\label{section:related work}

\subsubsection{Transfer Learning }Methods such as \cite{huang2006correcting, sugiyama2008direct, sun2011two, Shen2018WassersteinDG, lee2019sliced, duan2009domain, zhuang2009cross, evgeniou2005learning}, which focus on transferring the knowledge across domains, aim at improving the performance of target learners in target domains.  Compared with traditional transfer learning setting, our paradigm is not limited to 
static data, and has the ability to continuously recognize  unknown/new classes. Moreover, 
continual domain adaptation \cite{DBLP:conf/icml/ZhaoH10, DBLP:conf/nips/LiuLWW20, DBLP:journals/tkde/BitarafanBG16, DBLP:conf/icml/WangHK20, DBLP:conf/cvpr/ManciniBC019} enables the learner to adapt to continually evolving target domains without forgetting. However, this renders them impractical when deployed to the client because of model redundancy, while our method has no such restriction. Although meta domain adaptation \cite{DBLP:conf/nips/BalajiSC18, DBLP:conf/aaai/LiYSH18, Sun2019MetaTransferLF} uses a lightweight model, it cannot continually learn new 
claases in an open-world environment.

\subsubsection{Meta Learning }There is a three-way taxonomy across optimization-based methods \cite{DBLP:conf/icml/FinnAL17, DBLP:conf/nips/ShuXY0ZXM19}, model-based methods \cite{DBLP:conf/iclr/RaviL17, santoro2016meta}, and metric-based methods \cite{koch2015siamese, vinyals2016matching, sung2018learning, snell2017prototypical} in the meta-learning. They intend to design models as prior knowledge of learning model parameters that can learn new skills or adapt to new domain rapidly with few training examples. However, these approaches conventionally assumed that these tasks should be used together in batches, also called as few-shot learning, which are unable to train on sequential tasks. Although meta continual learning \cite{DBLP:conf/nips/YaoZMLSX20, gidaris2018dynamic, DBLP:conf/nips/JerfelGGH19, DBLP:conf/nips/JavedW19, DBLP:conf/nips/GuptaYP20} can solve the sequential task problem by recording the efficiency task buffer, it still cannot prevent data-interaction. The collective-individual paradigm may not be referred to as meta-learning, but 
it may harness meta-learning across learning tasks, referred to as learning to learn. Thus, we set up some comparative experiments in the Experimental Results section. 


\subsubsection{Continual Learning }Continual learning \cite{DBLP:journals/nn/ParisiKPKW19, de2019continual, kirkpatrick2017overcoming, DBLP:conf/nips/ShinLKK17, rusu2016progressive, rebuffi2017icarl, DBLP:conf/iclr/YoonYLH18, NEURIPS2018_cee63112} is also referred to as lifelong learning \cite{DBLP:series/synthesis/2018Chen, DBLP:conf/cvpr/AljundiCT17, DBLP:conf/iclr/ChaudhryRRE19}, which aims to continually learn over time by accommodating new knowledge while retaining previously learned experiences. However, when data sharing is restricted due to its proprietary or privacy issues, this reliance on the coexistence of source data and target data is very impractical. \emph{Open-world learning} (OWL, a.k.a. open world recognition or classification) \cite{DBLP:conf/cvpr/BendaleB15,DBLP:conf/icra/ManciniK0JC19,DBLP:journals/corr/abs-1801-05609} can be broadly defined as learning a model that can perform its intended task and then incrementally learn the new things. Furthermore, we solve the problem of transferring the meta-knowledge learned in the open-world to a practical lightweight model without data-interaction. Accordingly, a comparison of collective-individual paradigm with several different settings is provided in Table~\ref{tab:related-work}.
 
 To sum up, our formulation of the collective-individual paradigm is different from prior arts. Not only can it recognize unknown/unseen classes in the open-world, but it can also adapt to new specific tasks by reconstructing and initializing a lightweight model based on few data of novel classes.
 
\section{Methodology}
\subsection{Open-world Expandable/Collective Model}

\begin{figure}[ht]
\setlength{\abovecaptionskip}{0pt}
\setlength{\belowcaptionskip}{10pt}
\vskip -0.2in
\begin{center}
\centerline{\includegraphics[width=0.5\textwidth]{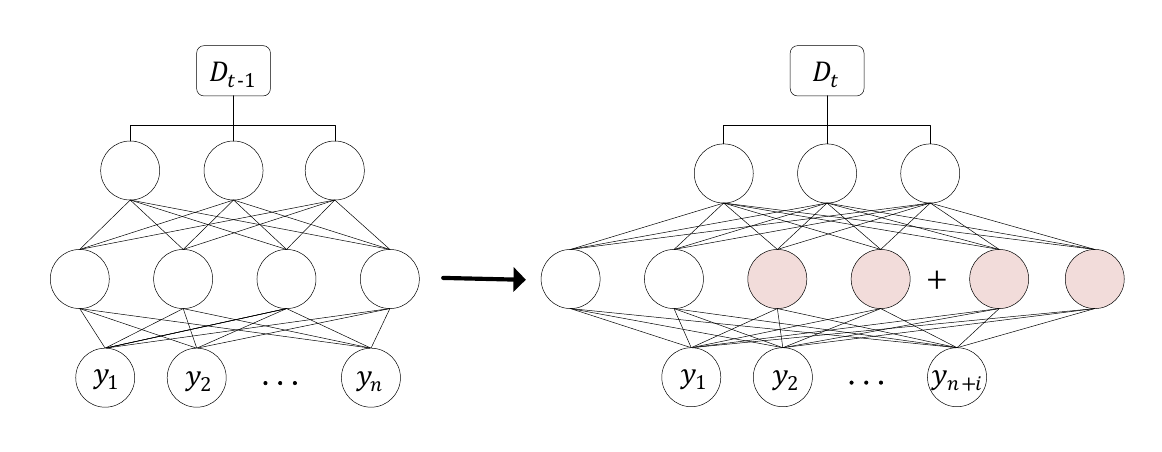}}
\caption{\emph{Expand the collective model.} Left: The network trained on the $\mathcal{D}_{t-1}$. Right: The new task $\mathcal{D}_{t}$ contains $i$ new classes. When training on $\mathcal{D}_{t}$, copy the model modules in the red part and splice them together. The collective model supports deepening and widening operations.   }

\label{fig:model-extension}
\end{center}
\vskip -0.4in
\end{figure}

\subsubsection{Model Expansion} 
 We presume that the model has a sequence of $T$ tasks, where the training data coming in at time $t$ is $D_t=\left\{\boldsymbol{x}_{i},y_{i}\right\}_{i=1}^{N_{t}}$. The tasks may contain new categories. Here we deliberate that each task is multi-class classification. When deepening the network, it arbitrarily duplicates the network of the previous layer to the following layer. Similarly, when the network is widened, a replica of the current layer can be randomly picked and then spliced together. Figure ~\ref{fig:model-extension} describe the expanding process of the collective model. The network at all times before $t$ is trained on the data of $\mathcal{D}_{1},\ldots,\mathcal{D}_{t-1}$. When dealing with $D_t$, the network capacity demands to be increased to hold more knowledge of new tasks. Subsequently, network parameters are relevant to $\mathcal{D}_{1},\ldots,\mathcal {D}_{t}$. We train the network with the subsequent objective function: 

\begin{equation}
\label{eq:minWF}
\operatorname{minimize}_{\boldsymbol{W}_{F}^{t}} \mathcal{L}\left(\boldsymbol{W}_{F}^{t} ; \boldsymbol{W}_{F}^{t-1}, \mathcal{D}_{t}\right),   t=1,\ldots,T,
\end{equation}

where $\mathcal{L}$ is a task-specific loss function(e.g., cross-entropy error function), $\boldsymbol{W}_{F}^{t}$ is the added parameter for task $t$, and $\boldsymbol{W}_{F}^{t-1}$ is the set of parameters copyed from the network at task $t-1$. We optimize the hyperparameters of the collective model by gradient descent. \footnote{See Supplementary A.1.1 for more details.}

\subsubsection{Open-world Recognition}
The collective model $\mathcal{M}_\mathrm{N}$ is trained on the tasks sampled from the base classes $\hat{\mathbf{Y}}^{\text {close }}$. More importantly, the collective model $\mathcal{M}_\mathrm{N}$ not only can discriminate a test instance belonging to the known $\mathrm{N}$ classes, but also recognize an unknown or unseen class instance by classifying it as an unknown. The unknown instance set $U$ can then be forwarded to a human user who can identify $\mathrm{M}$ new classes of interest (among a potentially large number of unknowns) and provide their training examples. The collective model incrementally adds $\mathrm{M}$ open-world classes and updates itself to generate a renewed model $\mathcal{M}_\mathrm{N+M}$, instead of learning from scratch on the full dataset. The known class set is also updated $C_{t+1} = C_t + \{\mathrm{N + 1},...,\mathrm{N + M}\}$. This cycle continues over the life of the collective model, where it adaptively shapes itself with new knowledge. For an unknown instance, we use the following formula to calculate its open-world probability:
\begin{equation}
\label{eq:probability}
\hat{\mathbf{P}}=1-\max \frac{exp(f(x, y_i;\theta_{col}))}{\sum_{j=1}^{N} exp(f(x, y_j; \theta_{col}))} .
\end{equation}
Based on $\hat{\mathbf{Y}}^{\text {close }}$ and $\hat{\mathbf{P}}$ , open-world pseudo label $\hat{\mathbf{Y}}^{\text {open }}$ is given as:
\begin{equation}
\label{eq:openclasses}
\hat{\mathbf{Y}}^{\text {open }}=\{\begin{array}{cl}C_{\text {open }} & \hat{\mathbf{P}} >\lambda_{\text {open }} \\ \hat{\mathbf{Y}}^{\text {close }} & \hat{\mathbf{P}} \leqslant \lambda_{\text {open }}\end{array}.
\end{equation}
where $C_{open}$ denotes the open-world class, and $\lambda_{open}$ is the threshold to determine open-world instance. Therefore, the collective model is supposed to recognize a unknown or unseen class as well as incrementally adds it.

\begin{figure}[ht]
\setlength{\abovecaptionskip}{0pt}
\setlength{\belowcaptionskip}{10pt}
\begin{center}
    \centerline{\includegraphics[width=0.55\textwidth,trim=1 1 1 50,clip]{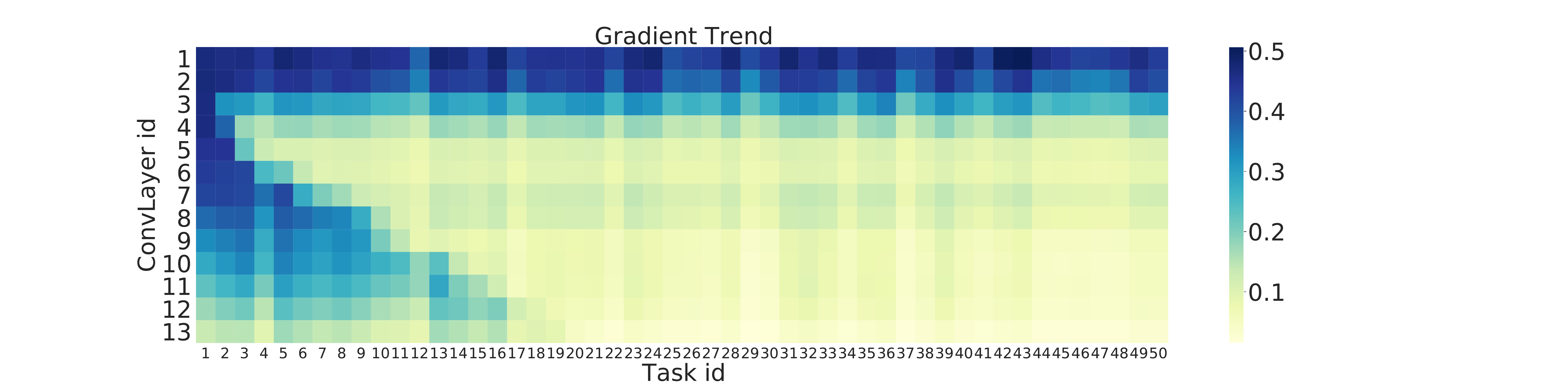}}
\caption{\emph{Gradient trends.} The collective model handles 50 sequential tasks sampling from ImageNet. It shows the proportion of the large gradient from each model layer according to the judgment criterion. The gradient of each layer in the collective model trained with the open-world tasks reflects the number of parameter changes. The deeper network parameter changes tend to be stable with the task order, which is judged as the expected learngene. }
\label{fig:gradient_task}
\end{center}
\vskip -0.4in
\end{figure}

\subsection{Inheriting Learngene that Represents Meta-knowledge}
In collective-individual paradigm, the primary challenge is to tackle how to determine the learngene by exploiting judgement criterion during training and the memorization effect of deep neural networks. Accordingly, learngene should have the ability to confirm critical modules and parameters across the tasks. Following this intuition, we design the criterion.

\subsubsection{Judgement Criterion}
For the optimization of the objective function $\mathcal{L}(\boldsymbol{W} ; \mathcal{D})$, the optimality will be achieved at $\boldsymbol{W}$ when $\nabla \mathcal{L}(\boldsymbol{W} ; \mathcal{D})=\mathbf{0}$ \cite{boyd2004convex, bubeck2014convex}. However, modern neural networks are complex and over-parameterized, which makes $\nabla \mathcal{L}(\boldsymbol{W} ; \mathcal{D})$ extremely high-dimensional. Optimality is difficult to judge effectively. To solve this problem, we use a more intuitive interpretation of the judgement criterion in this paper, so that optimization can be associated with a scalar. Every layer of our designed network is related to the data of $\mathcal{D}_{1},\ldots,\mathcal{D}_{t}$. Therefore, we analyze the gradient changes in the scalable neural network from the perspective of sequential tasks. Consider a parameter in one layer of the neural network, denoted by $\mathrm{w}_{i} \in \boldsymbol{W}$, its gradient is $\nabla \mathcal{L}\left(\mathrm{w}_{i} ; \mathcal{D}_{t}\right)$. Specifically, if we let 
\begin{equation}
\label{eq:gradient_scalar}
g_{i,t}=\left|\nabla \mathcal{L}\left(\mathrm{w}_{i} ; \mathcal{D}_{t}\right) \right|, i \in[m], t=1,\ldots
\end{equation}
where $m$ represents the number of parameter in a certain layer. In this way, the judgement can be checked by exploiting the scalar $g_{i,t}$. 

\subsubsection{Inherting Learngene}
We have shown that the optimization of the objective function can be related to a scalar $g_{i,t}$. There are three possible scenarios in open-world classification: (1) The parameters hardly change with the sequential task training, $g_{i,t}$ is always close to 0. (2) The parameters change drastically with the training of the sequence task, and $g_{i,t}$ is always large. (3) The gradient changes drastically and then stabilizes, that is, $g_{i,t}$ has a tendency to decrease and approach 0. Therefore, the judgement criterion is denoted by $\rho$, i.e., 
\begin{equation}
\label{eq:gradient_rate}
\rho = \frac{1}{m}\sum_{i=1}^{m}\Phi \left(g_{i,t}> \sigma\right),
\end{equation}
where $\Phi$ return 1 if $g_{i,t}$ exceeds the threshold $\sigma$, and 0 otherwise. $\rho$ represents the proportion of the large gradient of a certain convolutional layer. If the value of $\rho$ is from large to steadily small, the network layer is viewed as candidate learngene, as it corresponds to the third scenario. As shown by Figure~\ref{fig:gradient_task}, the parameter changes of deeper network layers tend to be stable with the task order, which are judged as the expected learngene. Furthermore, We hypothesize that, the top layer is more inclined to learn the semantic-level information among tasks, i.e., meta-knowledge, after training 
numerous tasks in time series. On the contrary, the parameters of the front(bottom) layers change drastically, which corresponds to the second scenario. Because the bottom layers are sensitive to category shift and specific tasks. \footnote{See Supplementary A.3.1 for more details. }

Although the collective model is expandable, the individual model that inherits learngene does not need a enormous network to adapt to the target tasks with few samples. Specifically, with the judgement criterion, we reuse the $topk$ layers of the network, i.e., the deepest $k$ layers as the critical task-agnostic modules and parameters. Subsequently, the learngene is connected to the lightweight network layers for reconstructing the individual model. 

\subsection{Adapt Novel Classes with Few Samples}

\begin{algorithm}[tb]
   \caption{Adapt from the Collective Model}
   \label{alg:heritance}
\begin{algorithmic}
   \STATE {\bfseries Input:} feature layers $F_{col}$ in the collective model, lightweight feature layers $F_{ind}$ in the individual model, classifier $C$ in the individual model, number of samples $\eta$ \\

   $F_{l} \leftarrow$ pick $topk$ layers in $F_{col}$\\
   individual model $\mathcal{M}$ $\leftarrow$ $\text{RECONSTRUCT}(F_{ind}, F_{l}, C)$\\
   parameter $\theta_{\mathcal{M}}$ of the model $\mathcal{M}$, weight factor $\lambda$
   \FOR{$i=1,\ldots,n$}
   \WHILE{$\mathcal{L}_{rec}+{\lambda}\mathcal{L}_{ret}$ has not converged}
   \STATE $(X_{c}, Y_{c}) \leftarrow$ batch sampled from $Q_{i}$
   \STATE $\widehat{Y}_{c}\leftarrow \mathcal{M}\left(X_c\right)$
   \STATE compute $\mathcal{L}_{rec}$ using Eq. ~\ref{eq:adploss}
   \STATE $X_{p}$ batch sampled from $Q_{i}$
   \FOR{$x_{p}$ in $X_{p}$}
   \STATE $\mathbf{L}\leftarrow$ $log\left[\sigma\left(\mathcal{M}(x_p)\right)\right]$
   \STATE fisher matrix $F_{x}$ $\leftarrow\frac{\partial \mathbf{L}}{\partial x_{p}} $ 
   \ENDFOR
   \STATE Compute $\mathcal{L}_{ret}$ using Eq. ~\ref{eq:genloss}
   \STATE update $\theta_{\mathcal{M}}$ by minimizing Eq. ~\ref{eq:toltalloss}
   \ENDWHILE 
   \ENDFOR
  
\end{algorithmic}
\end{algorithm}

\begin{figure*}[!htb]
\setlength{\abovecaptionskip}{0pt}
\setlength{\belowcaptionskip}{0pt}
  \centering
    \subfigure[Performance comparison on CIFAR100 ]{
    \label{fig:Cifar100-accuracy}
    \includegraphics[width=0.22\textwidth,trim=1 1 1 20,clip]{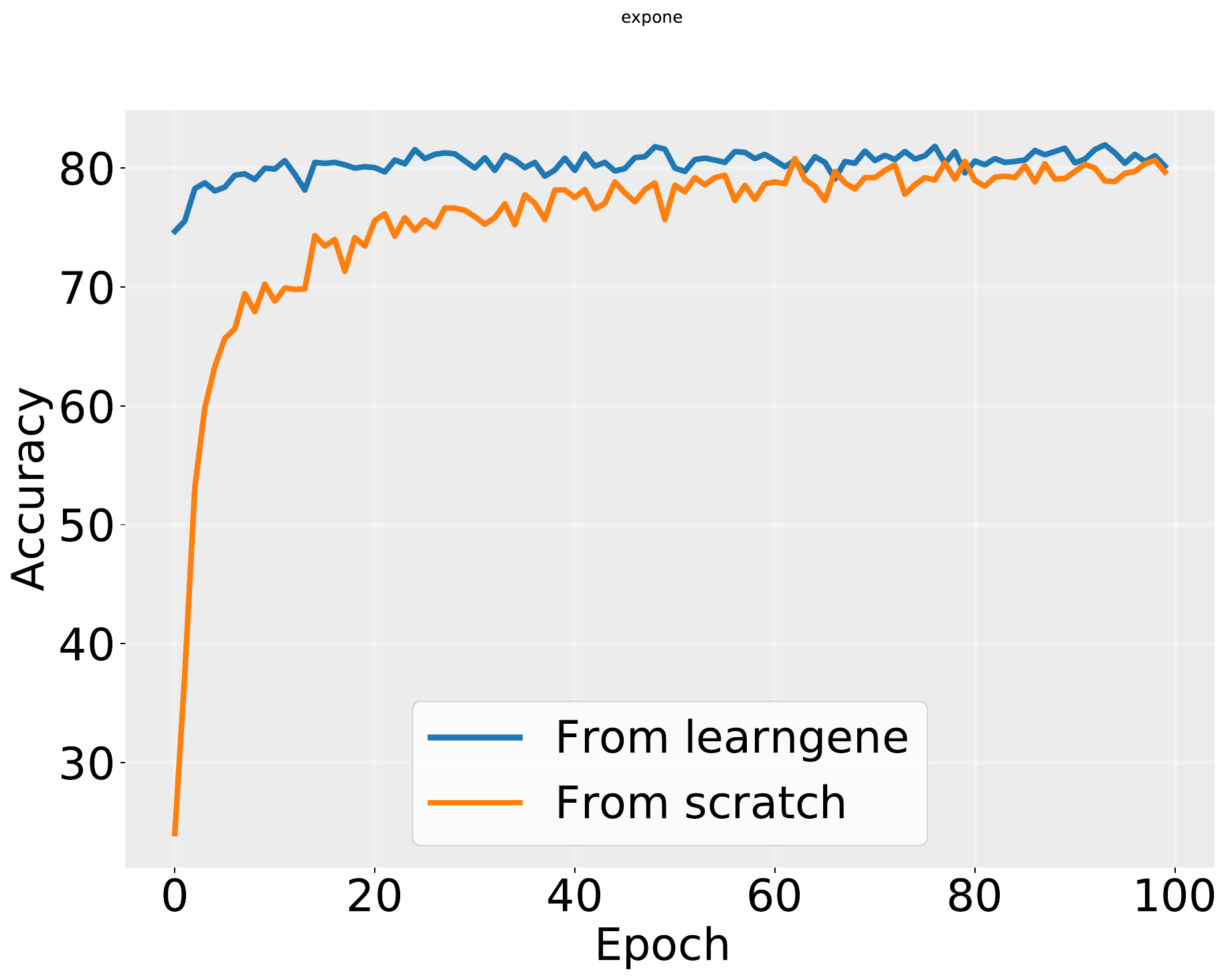}} 
    \subfigure[Performance comparison on ImageNet100 ]{
    \label{fig:ImageNet100-accuracy}
    \includegraphics[width=0.22\textwidth,trim=1 1 1 20,clip]{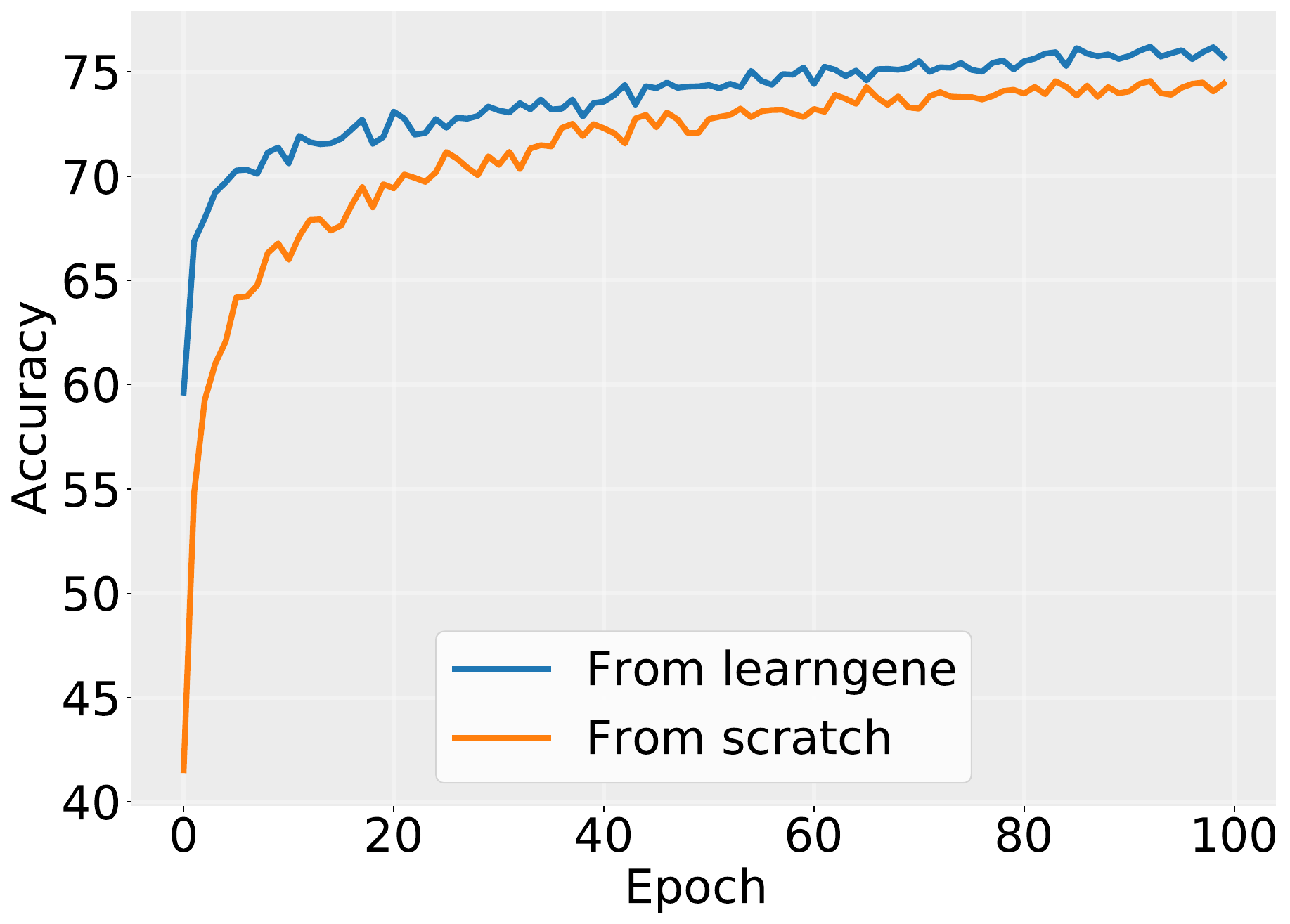}} 
    \subfigure[Effect of number of samples on CIFAR100.]{
    \label{fig:Cifar100-samples}
    \includegraphics[width=0.22\textwidth,trim=1 1 1 20,clip]{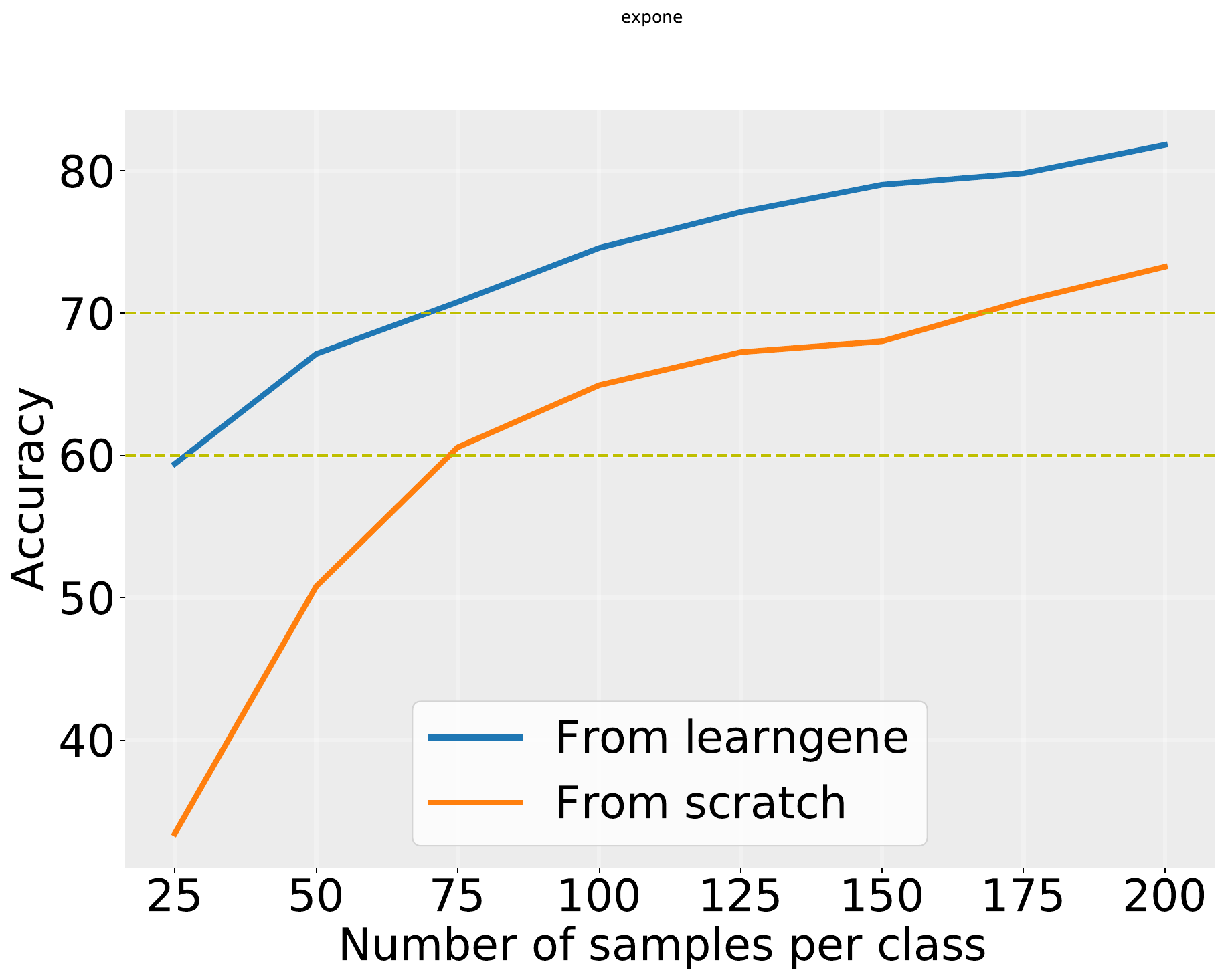}} 
    \subfigure[Effect of number of samples on ImageNet100.]{
    \label{fig:ImageNet100-samples}
    \includegraphics[width=0.22\textwidth,trim=1 1 1 20,clip]{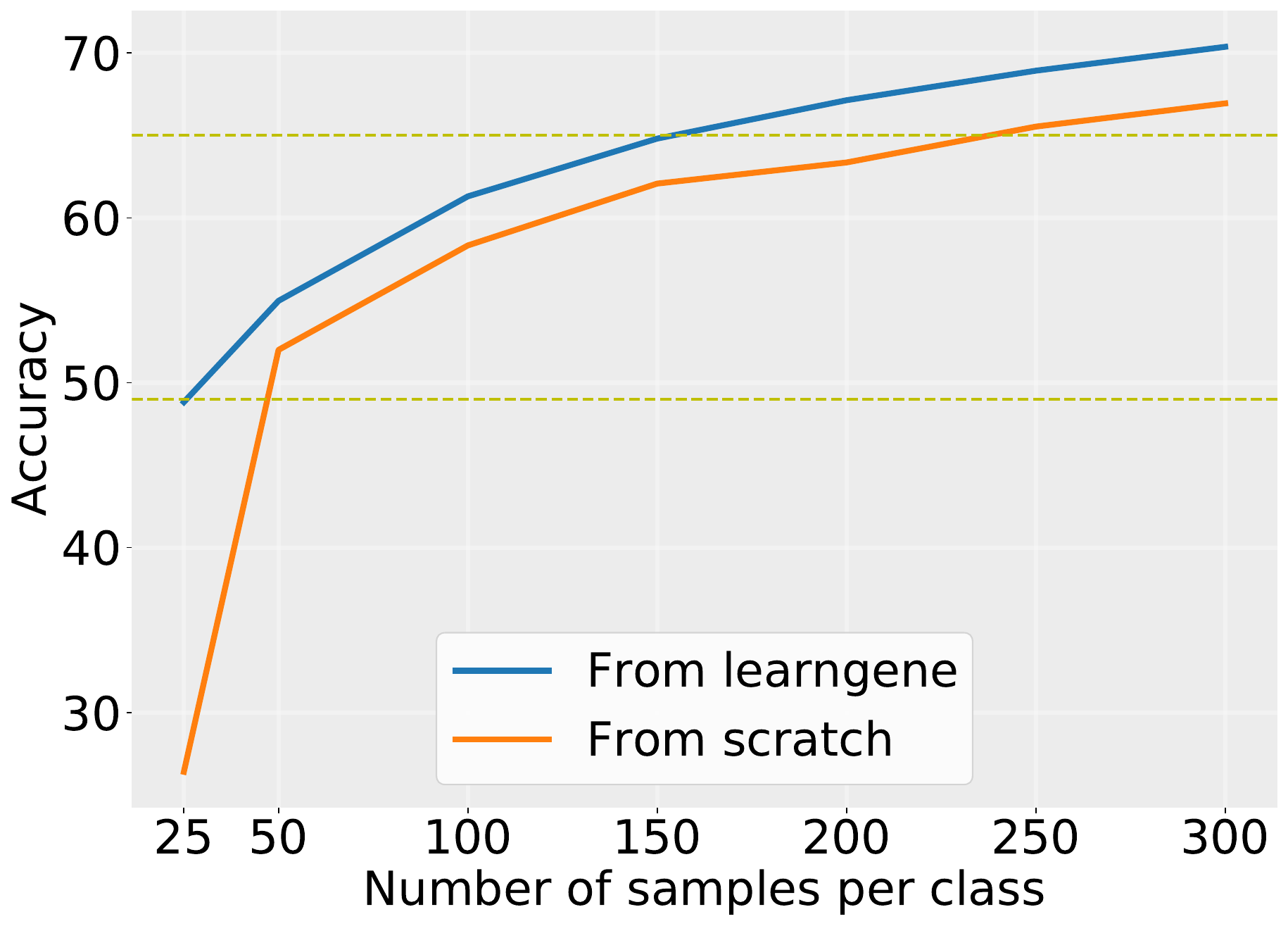}} 
  \caption{\emph{Comparison with learning from scratch. }Figure (a) and (b) reveal that the individual model reconstructed by learngene reached a result close to convergence with averaged over 5 and 10 runs, and it is consistently higher than the one learned from scratch. Figure (c) and (d) show that the individual model requires far fewer samples to achieve fixed accuracy and perform much better with fewer samples than learning from scratch. }
    \label{fig:R1}
    \vspace{-0.1in}
\end{figure*}

Due to the meta-knowledge of learngene from the collective model, the individual model greatly reduces the number of training samples. The adaptation contains two processes (see Algorithm~\ref{alg:heritance}): reconstruct and retain.

\subsubsection{Reconstruct}
The stage of adaptation allows the model to effectively respond to specific tasks of novel classes and reconstruct individual model based on learngene. We reuse the $topk$ modules and parameters of the collective model, which are connected to the smaller network layers. Therefore, the loss function of reconstruction is formulated as follows:
\begin{equation}
\label{eq:adploss}
\mathcal{L}_{rec}= L_{\mathrm{ce}}(f(x;\theta_{ind}), \mathcal{Y}),
\end{equation}
where $L_{\mathrm{ce}}$ is cross entropy loss.

\subsubsection{Retain}
The stage of inherting learngene decides to retain knowledge from the collective model. Retaining information of learngene can help the individual model to pass the early stage smoothly. At the same time, it is obligatory to maintain a certain degree of flexibility to adapt the network to new tasks. \cite{zador2019critique} also emphasizes a similar balance between innate knowledge and acquired learning. Consequently, we regularize  
reused learngene to suppress the update of the parameters. Fisher information \cite{kirkpatrick2017overcoming} roughly measures the sensitivity of the model output distribution to small changes in parameters. For the supervised learning model $p_{\theta}\left(y|x\right)$, the fisher information matrix for a specific input $x$ is defined as follows:
\begin{equation}
F_{x}=\mathbb{E}_{p_{\theta}(y \mid x)}\left[\nabla_{\theta} \log p_{\theta}(y \mid x) \nabla_{\theta} \log p_{\theta}(y \mid x)^{\top}\right].
\end{equation}
Therefore, using Fisher information as a secondary penalty, it penalizes the change of parameter distribution (measured by KL divergence). Limiting the slight changes in network parameters leads to the actual functional changes of the model also small. Given the fisher information matrix, the loss function of retaining $\mathcal{L}_{ret}$ is defined as
\begin{equation}
\label{eq:genloss}
\mathcal{L}_{ret}=\sum_{i=n-k+1}^{n} \frac{1}{2}\left(F_{i}\left(\theta_{i}-\theta_{A, i}^{*}\right)^{2}\right).
\end{equation}
 $k$ means that only the $topk$ network layers are suppressed in the individual model, which is learngene inherited from the collective model. 
Finally, the loss function of the individual model is
\begin{equation}
\label{eq:toltalloss}
\mathcal{L}=\mathcal{L}_{rec}+{\lambda}\mathcal{L}_{ret},
\end{equation}
where $\lambda$ is the weighting parameter to trade-off between reconstruction and retaining.

\subsection{Theoretical Analysis}
Here, we present the theoretical analysis for transfering representation from collective model and get the following bound based on PAC-Bayesian \cite{DBLP:conf/icml/PentinaL14}\\
\textbf{Theorem 1. } \emph{For the prior $P$, we choose a Gaussian with zero mean and variance $\sigma I_{k}$. The posterior $Q$, is a shifted Gaussian with variance $\sigma I_{k}$ and mean $w_{Q}$ in the same subspace. The following bound holds for $t$ tasks }
\begin{equation}
\label{eq:theorem-2}
er(h)-\widehat{er}(h)<\frac{c_{2}}{2 \sigma t \sqrt{\bar{m}}}+\mathit{c_{1}} ,
\end{equation}

where $c_{1}$ and $c_{2}$ are all constants. $\bar{m}=\frac{1}{n} \sum_{i=1}^{n} {m_{i}}$ is the mean of the sample sizes. By inherting meta-knowledge from the collective model, we condition the individual model to satisfy Eq. \ref{eq:theorem-2}, which is considered promising for target learning tasks. Since we use learngene to transfer representation from the collective model and initialize the individual model, the convergence rate of generalization error is bound. Note that, with increasing number of labeled target instances (increasing $\bar{m}$), the first term in Eq. 10 decreases. It explains that fewer samples are needed in the individual model to achieve better performance. In our formulation, this is achieved by enforcing $L_{rec}$, which can be regarded as a way to self-supervise the individual model. \footnote{The proof is given in supplementary A.2}
\section{Experiments}
\subsection{Experimental setting}

\textbf{Datasets. } \textbf{1) CIFAR100.} This dataset consists of 60, 000 images of 100 generic object classes \cite{krizhevsky2009learning}. The collective model uses 60 classes as base classes, 16 classes as open-world classes, and the remaining 20 classes as novel classes, to ensure no data-interaction between the two models. Five classes are randomly sampled for each continual learning task. \textbf{2) ImageNet100. } This dataset has 14,197,122 images of 21841 labels \cite{DBLP:journals/corr/RussakovskyDSKSMHKKBBF14}. Since the full ImageNet dataset is very large, we sub-sample
100 classes  and conduct all experiments on this subset. Similarly, ImageNet100 is also divided into three parts and  resizes
84 × 84 for each class. \footnote{The details of data division are given in supplementary A.1.2}

\textbf{Basic network. } 1)\emph{ Collective model.} For experiments on the CIFAR100 and Imagenet100 datasets, we use thirteen convolutional layers as base network. The task data used by the collective model is inputted in chronological order.
2)\emph{ Individual model. }Considering the individual model is lightweight, we utilize seven convolutional layers, which the last three convolutional layers are inherited from the collective model. This individual model is adapted separately to target tasks of novel classes. \footnote{See Supplementary A.1.3 for more details}

\textbf{Hyperparameters. }
We set the learning rate to 0.001 and 0.005 when training the collective model and individual model, respectively. We fix all batch size to 32 and sample size in estimating fisher information to 128. $\lambda_{out}$ is 0.9587 for CIFAR100 and 0.9733 for Imagenet100 to achieve the best performance.

\subsection{Experimental results and analysis}
\label{sec:experimentalresults}
\textbf{Evaluation on benchmark datasets. }
We report results on CIFAR100 and ImageNet100 datasets in Figure ~\ref{fig:R1} (a) and (b). Learning curves in Figure ~\ref{fig:R1} (a) show that the individual model is the fastest to adapt to new tasks in 10 epochs, averaging five target tasks. By contrast, Learning from scratch requires 40 epochs to converge. 
We also find that the individual model substantially outperforms the alternative approaches most of the time, confirming the useful knowledge inherited from the collective model. We hypothesize that, the learngene improve in efficiency over the course of learning as they see tasks previously. Besides, learngene has the ability to pay attention to meta-knowledge among tasks and generalize to novel classes. On the contrary, learning from scratch does not learn quickly compared to the individual model because there is no effective initialization rules. 


\textbf{Effect of number of training samples. }
We evaluate the effectiveness of the individual model from the perspective of samples. The results on CIFAR100 and ImageNet100 datasets are reported in Figure ~\ref{fig:R1} (c) and (d). As shown in Figure ~\ref{fig:R1} (c), the individual model only fits 25 samples to achieve 60\% performance after 10 epochs. By contrast, Learning from scratch is overfitting such a small sample, and the gap is very obvious, where it attains at least 20\% lower average accuracy. In addition, in abundant samples, the individual model has always performed better than learning from scratch. On the other hand, compared with learning from scratch, the individual model saves 2/3 of the sample and fine-tuning to 60\% accuracy after 10 epochs. Furthermore, when the effect reaches 70\% average accuracy, the individual model saves 100 samples for each class and extremely reduces the training samples for model. Therefore, it is possible to deploy the model to the personal user terminal. 


\begin{table}[h]
\setlength{\abovecaptionskip}{0pt}
\setlength{\belowcaptionskip}{0pt}
\caption{Multi-class classification accuracies on CIFAR100 and ImageNet100 with data of open-world classes.}
\label{tab:related-work}
\begin{center}
\begin{scriptsize}
\begin{sc}
\begin{tabular}{ccc}
\hline
Method           & CIFAR100       & ImageNet100    \\ \hline
MAML             & 34.07$\pm$0.63 & 36.51$\pm$1.01 \\
first-order MAML & 25.40$\pm$0.50 & 32.82$\pm$0.65 \\
PROTONET         & 33.20$\pm$0.48 & 36.26$\pm$0.32 \\
PROTO-MAML       & 39.09$\pm$0.35 & 32.57$\pm$1.28 \\
MATCHINGNET      & 34.99$\pm$0.34 & 37.79$\pm$0.39 \\
Fine-tuning      & 38.72$\pm$1.13 & 40.72$\pm$1.36 \\
Fine-tuning++    & 38.83$\pm$1.51 & 37.08$\pm$1.04 \\
DEEPEMD          & 39.85$\pm$0.41 & 40.21$\pm$0.82 \\ \hline
Ours             & \textbf{45.88}$\pm$2.99 & \textbf{42.40}$\pm$2.35 \\ \hline
\end{tabular}
\vspace{-0.1in}
\end{sc}
\end{scriptsize}
\end{center}
\end{table}

\textbf{Open-world and novel classes with few samples. }
As aforementioned in the related work, we also conduct a comparative experiment on the data of open-world and novel classes. Tables 2 and 3 show the performance of our model from learngene against the few-shot learning algorithm with architectures of seven convolutional layers. Values represent average five classification accuracies obtained from the data of novel classes in 30 epochs and tested directly on the data of open-world classes. In order to ensure a fair comparison, we reset the capacity of the architectures of eight representative methods (\cite{DBLP:conf/icml/FinnAL17, DBLP:journals/corr/abs-1803-02999, snell2017prototypical, Triantafillou2020Meta-Dataset, vinyals2016matching, Zhang2020DeepEMDFI}, two fine-tuning methods of \cite{DBLP:conf/iclr/ChenLKWH19}) to match ours on the data of novel classes. Our method can recognize unknown/unseen classes in the open-world scenario and incrementally add them. Still, the few-shot learning algorithm (trained on 20-shot) does not have this ability and can only directly test the open-world classes. As shown in the table 2, the collective model shows competitive results with open-world classes, achieving the best performance for all sample settings. For example, the individual model improves on average of a relative 6.03\% wrt DeepEMD (the second-best method) with the data of open-world classes on CIFAR100. Despite its simplicity, table 3 shows that our proposed method achieves an average accuracy that, on CIFAR100 and ImageNet100 with data of novel classes, is superior to state of the art with the same architectures. 

\begin{table}[h]
\setlength{\abovecaptionskip}{0pt}
\setlength{\belowcaptionskip}{0pt}
\caption{Multi-class classification accuracies on CIFAR100 and ImageNet100 with data of novel classes.}
\label{tab:related-work}
\begin{center}
\begin{scriptsize}
\begin{sc}
\setlength{\tabcolsep}{1mm}
\begin{tabular}{ccccc}
\hline
                     & \multicolumn{2}{c}{CIFAR100, 5way} & \multicolumn{2}{c}{ImageNet100, 5way}                                  \\
Method               & 10-shot          & 20-shot          & \multicolumn{1}{l}{10-shot}        & \multicolumn{1}{l}{20-shot}        \\ \hline
MAML                 & 51.70$\pm$1.75   & 62.36$\pm$2.88   & 48.62$\pm$3.10                     & 57.62$\pm$2.42                     \\
first-order MAML     & 50.13$\pm$3.73   & 60.00$\pm$3.31   & 43.38$\pm$2.47                     & 51.87$\pm$1.96                     \\
PROTONET             & 56.40$\pm$2.88   & 61.02$\pm$3.18   & 49.91$\pm$3.36                     & 58.62$\pm$2.45                     \\
PROTO-MAML           & 53.11$\pm$1.25   & 59.57$\pm$1.34   & 45.14$\pm$2.52                     & 53.57$\pm$1.34                     \\
MATCHINGNET          & 56.71$\pm$4.48   & 60.98$\pm$3.09   & 51.82$\pm$2.58                     & 59.78$\pm$2.94                     \\
Fine-tuning          & 48.98$\pm$3.86   & 60.89$\pm$3.77   & 51.11$\pm$3.36                     & \textbf{60.71}$\pm$2.32            \\
Fine-tuning++        & 52.67$\pm$4.02   & 61.20$\pm$3.29   & 40.80$\pm$2.92                     & 51.78$\pm$2.81                     \\
DEEPEMD              & \textbf{58.03}$\pm$1.85   & 63.69$\pm$2.05   & 50.93$\pm$2.78                     & 59.69$\pm$2.47                     \\ \hline
Ours($\lambda=1000$) & \textbf{58.86}$\pm$3.12   & \textbf{64.24}$\pm$1.89   & \multicolumn{1}{l}{\textbf{52.96}$\pm$4.84} & \multicolumn{1}{l}{\textbf{60.86}$\pm$4.48} \\
Ours($\lambda=5000$) & 58.21$\pm$2.58   & 63.74$\pm$2.46   & \multicolumn{1}{l}{52.53$\pm$4.74} & \multicolumn{1}{l}{59.76$\pm$5.41} \\ \hline
\end{tabular}
\vspace{-0.2in}
\end{sc}
\end{scriptsize}
\end{center}
\end{table}

\subsection{Discussion}

\begin{figure}[h]
    \vspace{-0.1in}
    \subfigure[Evolution (Expansion) of the collective model]{\label{fig:evolution}
    \includegraphics[width=0.22\textwidth,trim=1 1 1 20,clip]{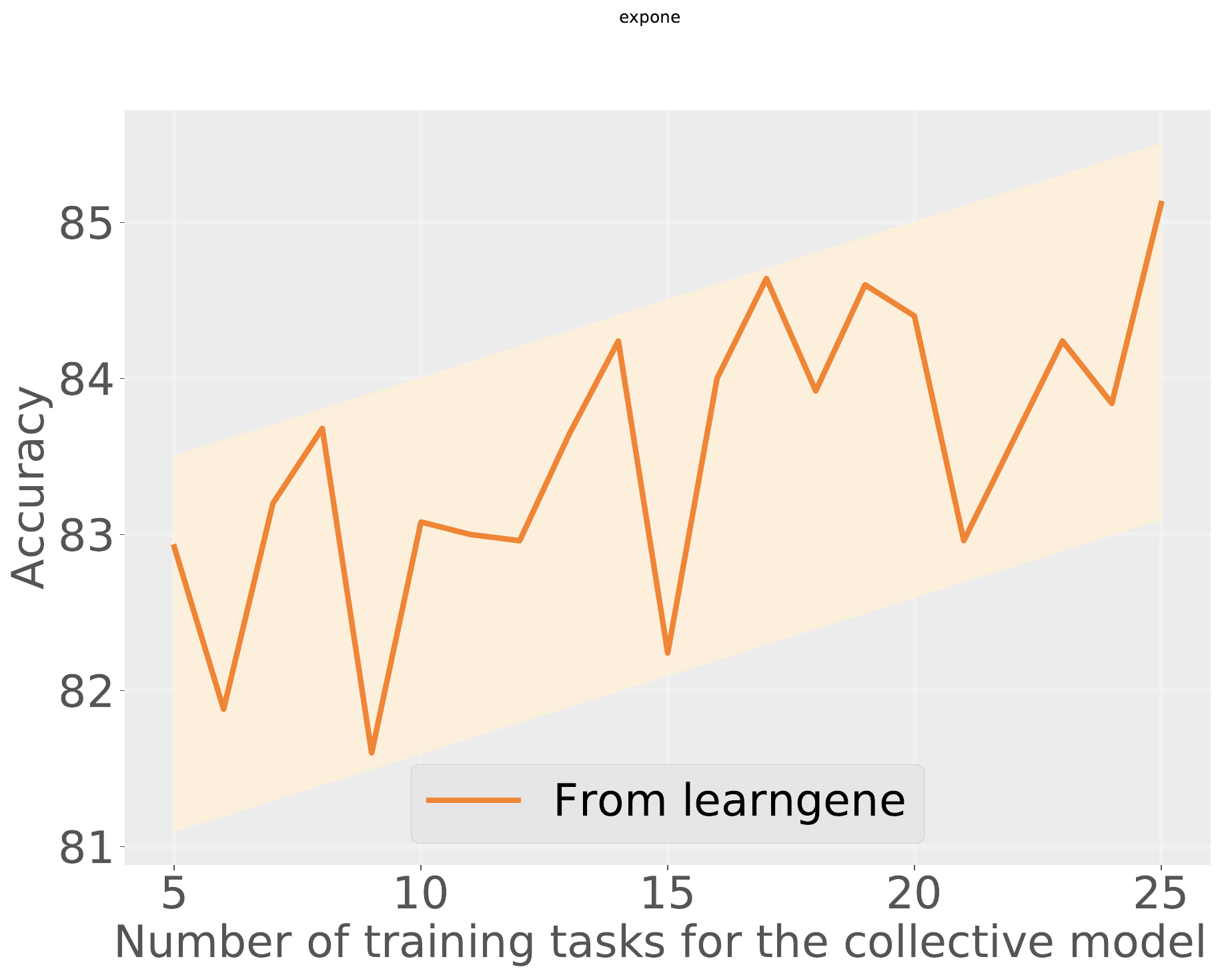}}
    \subfigure[Evaluation on judgement criterion]{
    \label{fig:necessity}
    \includegraphics[width=0.22\textwidth,trim=1 1 1 20,clip]{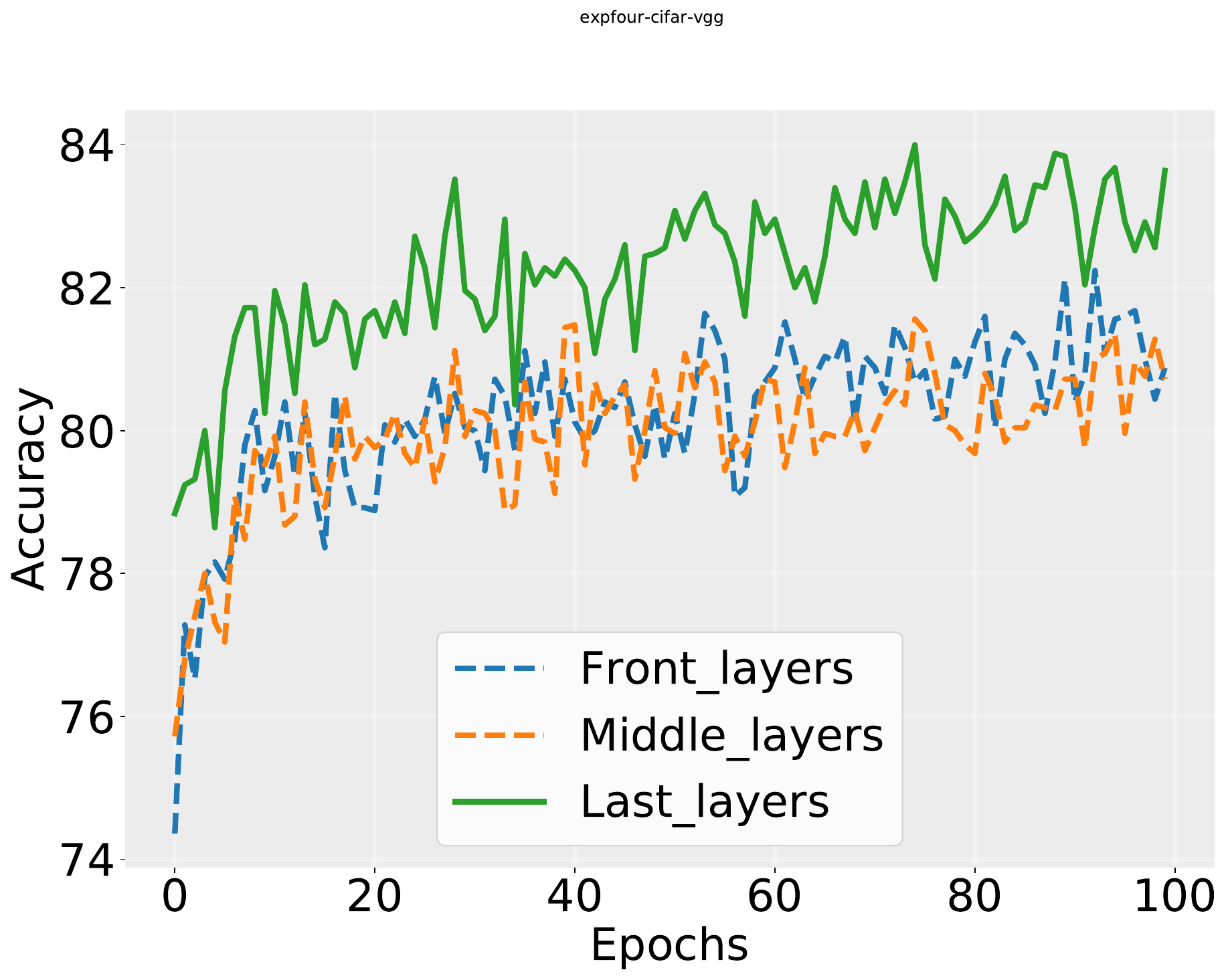}}
    \caption{Figure (a) reveals that the performance of the individual model continually improves, as the collective model is trained on sequence tasks. Figure (b) further verifies the rationality of judgement criterion for obtaining learngene.}
    \label{fig:R2}
    \vspace{-0.2in}
\end{figure}

\textbf{Evolution process of collective model. }
Intuitively, the collective model evolves (expands) \cite{stearns2000evolution} on sequential tasks, and then the individual model reconstructed from learngene performs better over time. Inspired by this, we design an experiment to reconstruct multiple individual models with learngene, which is obtained from collective model trained along chronological order. The results (see Figure ~\ref{fig:R2}(a))  demonstrate that the individual model tends to perform better as the collective model continues to evolve (expand), since it alleviates catastrophic forgetting in the collective model and generate better meta-knowledge or semantic information. In the next subsection, we use visualization to explain this process.

\begin{figure}[ht]
\setlength{\abovecaptionskip}{0pt}
\setlength{\belowcaptionskip}{10pt}
\vspace{-0.1in} 
\begin{center}
\centerline{\includegraphics[width=0.55 \textwidth,trim=200 140 200 100,clip]{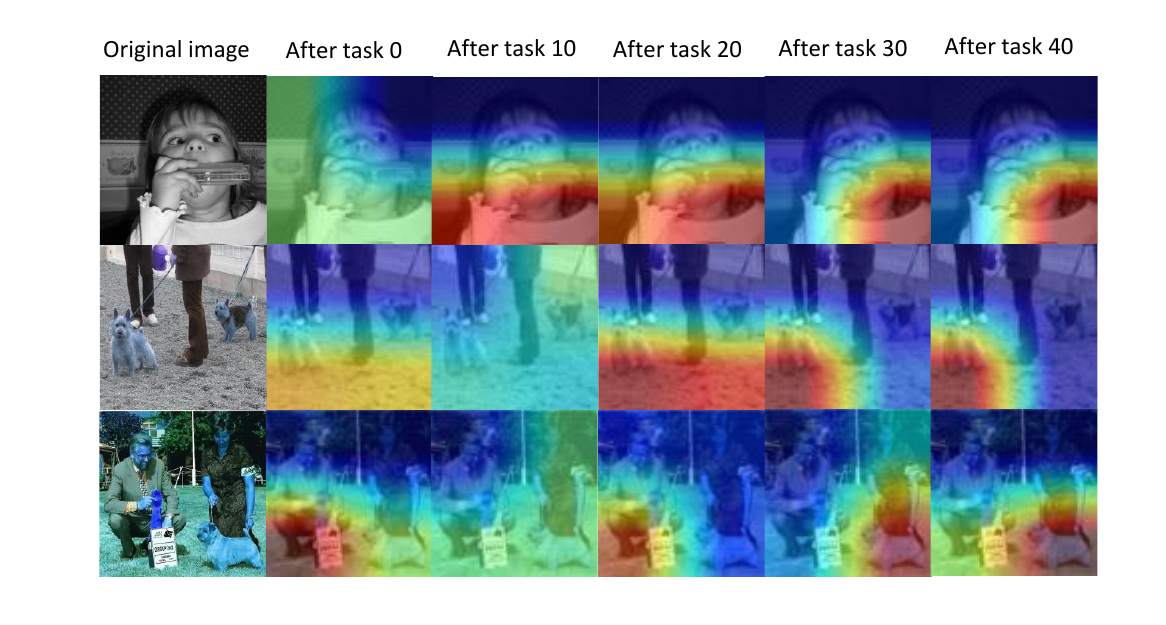}}
\caption{\textbf{Visualization for the collective model on unknown/unseen ImageNet samples.} This figure analyzes that the collective model pays more attention the important position of the sample with sequential tasks training. For example, the first row reveals that the collective model gradually focuses on the harmonica in the lower right corner, when facial features and one hand interfere. Since the second and third rows of scenes are very similar, the model ultimately focuses on a similar semantic, such as the terrier.}
\label{fig:visualization}
\end{center}
\vspace{-0.4in} 
\end{figure}

\textbf{Why is learngene needed? } We further discuss the interpretability of learngene through visualization  \cite{DBLP:conf/iccv/SelvarajuCDVPB17}, derived from the output of top convolutional layer in the collective model. We use the other classes in ImageNet to visualize in the training process of the collective model. Since the collective model is continuously trained, Figure ~\ref{fig:visualization}  shows that the model pays more attention to the position of the similar semantic information for the unseen classes. Therefore, learngene inherited from the collective model also has the ability to aid the individual model in quickly distinguishing the unseen categories, which also explains why unknown instances can be recognized.

\textbf{Evaluation on judgement criterion. }
In order to further confirm the learngene proposed by our judgement criterion, we design a network with the same capacity for each layer. The network is also trained on sequence tasks. Thereafter, we squeeze the front, middle and top three areas of the network to reconstruct the individual model. Figure ~\ref{fig:R2}(b) shows that the individual model inheriting learngene from the top position of the network is stable and finally 2\% accuracy higher than other locations. Accordingly, since the top network layers can better capture semantic information or meta-knowledge between tasks, the individual model reconstructed by learngene can adapt to the target tasks more quickly and has better generalization performance.

\section{Conclusion}
In this paper, we propose a practical collective-individual paradigm. Furthermore, we introduce the learngene that inherits the meta-knowledge from the collective model in the open-world scenario and reconstructs a new lightweight model for the target task. We use a novel and effective criterion to discover learngene based on gradient information. Through extensive empirical evaluation and theoretical analysis, we demonstrate the effectiveness of our approach. As future work, networks of different structures (e.g. Resnet \cite{DBLP:conf/cvpr/HeZRS16}) and automatically inherit learngene as the initialization rule of the target task are worth exploration.

\section{Acknowledgments}
We sincerely thank Tiankai Hang and Jun Shu for helpful discussion. This research was supported by the National Key Research and Development Plan of China (No. 2018AAA0100104), and the National Science Foundation of China (62125602, 62076063). Moreover, We thank MindSpore for the partial support of this work, which is a new deep learning
computing framework\footnote{https://www.mindspore.cn/}.

\bibliography{aaai22}

\end{document}